\documentclass[letterpaper]{article} % DO NOT CHANGE THIS
\usepackage{aaai2026}  % DO NOT CHANGE THIS
\usepackage{times}  % DO NOT CHANGE THIS
\usepackage{helvet}  % DO NOT CHANGE THIS
\usepackage{courier}  % DO NOT CHANGE THIS
\usepackage[hyphens]{url}  % DO NOT CHANGE THIS
\usepackage{colortbl}
\usepackage{multirow}
\usepackage{amsmath}
\usepackage[table]{xcolor}
\usepackage{graphicx} % DO NOT CHANGE THIS
\usepackage{natbib}  % DO NOT CHANGE THIS AND DO NOT ADD ANY OPTIONS TO IT
\usepackage{caption} % DO NOT CHANGE THIS AND DO NOT ADD ANY OPTIONS TO IT
\usepackage{algorithm}
\usepackage{algorithmic}
\usepackage{booktabs}    % 美化横线

\usepackage{newfloat}
\usepackage{listings}
\DeclareCaptionStyle{ruled}{labelfont=normalfont,labelsep=colon,strut=off} % DO NOT CHANGE THIS
\floatstyle{ruled}
\newfloat{listing}{tb}{lst}{}
\floatname{listing}{Listing}
\definecolor{mygray}{gray}{0.75}
\definecolor{myred}{rgb}{1, 0.75, 0.75}  

\title{DANS-KGC: Diffusion Based Adaptive Negative Sampling for Knowledge Graph Completion}
\author{
Haoning Li,
Qinghua Huang\thanks{Corresponding Author}
}
\affiliations{
   School of Artificial Intelligence, OPtics and ElectroNics (iOPEN), Northwestern Polytechnical University\\
    1134067491@qq.com, qhhuang@nwpu.edu.cn
}
\begin{document}

\maketitle

\begin{abstract}

Negative sampling (NS) strategies play a crucial role in knowledge graph representation. In order to overcome the limitations of existing negative sampling strategies, such as vulnerability to false negatives, limited generalization, and lack of control over sample hardness, we propose DANS-KGC (Diffusion-based Adaptive Negative Sampling for Knowledge Graph Completion). DANS-KGC comprises three key components: the Difficulty Assessment Module (DAM), the Adaptive Negative Sampling Module (ANS), and the Dynamic Training Mechanism (DTM). DAM evaluates the learning difficulty of entities by integrating semantic and structural features. Based on this assessment, ANS employs a conditional diffusion model with difficulty-aware noise scheduling, leveraging semantic and neighborhood information during the denoising phase to generate negative samples of diverse hardness. DTM further enhances learning by dynamically adjusting the hardness distribution of negative samples throughout training, enabling a curriculum-style progression from easy to hard examples. Extensive experiments on six benchmark datasets demonstrate the effectiveness and generalization ability of DANS-KGC, with the method achieving state-of-the-art results on all three evaluation metrics for the UMLS and YAGO3-10 datasets.

\end{abstract}

% Uncomment the following to link to your code, datasets, an extended version or similar.
% You must keep this block between (not within) the abstract and the main body of the paper.
% \begin{links}
%     \link{Code}{https://aaai.org/example/code}
%     \link{Datasets}{https://aaai.org/example/datasets}
%     \link{Extended version}{https://aaai.org/example/extended-version}
% \end{links}

\section{Introduction}

%As the core vehicle for structured knowledge representation, knowledge graph (KG) is essentially a semantic-network system formed by a multi-relational graph structure \cite{ref1}. By encoding real-world ontological knowledge in a standardized triple format: $\langle h, r, t \rangle$, where $h$, $r$, and $t$ represent head entity, relation, and tail entity, respectively.  Knowledge graphs have been widely adopted in AI domains such as recommender systems, information retrieval, and intelligent question answering, demonstrating notable engineering value and theoretical significance. Nevertheless, because of data-collection limitations and the dynamic evolution of knowledge, existing knowledge graphs often exhibit missing entities and relations. To tackle this challenge, knowledge-graph completion (KGC) techniques have emerged, dedicated to uncovering latent semantic associations among entities and thereby systematically improving the structural completeness of knowledge graphs \cite{ref2}.

Knowledge graphs (KGs) encode real‑world facts as triples $\langle h,r,t \rangle$ in a multi‑relational graph, serving as a key foundation for recommender systems, search, and QA \cite{ref1}. Yet data limitations and continual change leave many entities or relations missing. Knowledge‑graph completion (KGC) therefore seeks to infer hidden semantic links and restore KG integrity \cite{ref2}.

After years of research, a rich spectrum of KGC techniques has been proposed and widely deployed across diverse domains, consistently delivering strong empirical performance. Current approaches can be roughly classified into: translation-based methods, rotation-based methods, semantic matching-based methods, and neural network-based methods \cite{ref3}. Although these methods differ in how they represent triples, most of them adopt a common training paradigm centered on positive\&negative sample pairs. Positive samples correspond to factual triples in KG, whereas negative samples are constructed via negative sampling to act as semantic foils. During training, the model learns by pulling positive pairs closer together in the embedding space while pushing negative pairs farther apart, thereby capturing the decision boundary and relational semantics. Consequently, the negative sampling (NS) strategy is pivotal in KGC, the quality of generated negative pairs directly determines the fidelity of triple embeddings and, by extension, the performance of downstream tasks such as link prediction and multi-hop reasoning \cite{ref4}.

Negative sampling is one of the core components determining model performance in KGC. Its goal is to construct high-quality contrast triples (negative samples) for factual triples (positive samples), enabling the model to more clearly learn semantic boundaries and reasoning patterns \cite{ref5}. Traditional negative sampling strategies include random replacement \cite{ref6}, heuristic rule–based approaches \cite{ref7}, and adversarial negative sampling \cite{ref8}. While these methods have progressively improved the quality of negative samples in knowledge graph completion, they still face limitations such as susceptibility to false negatives, poor generalizability, and difficulty in controlling the hardness distribution of negative samples \cite{ref9}. With the remarkable success of diffusion models in generative tasks, diffusion-based negative sampling strategies have emerged \cite{ref10}. These methods generate samples through a step-wise denoising Markov chain: early timesteps produce coarse, easy-to-distinguish negative samples, while later steps generate semantically closer and harder negatives \cite{ref11}. This naturally forms a distribution of negative samples ranging from easy to hard. The approach not only enhances diversity but also offers controllability.

Although diffusion-based negative sampling can naturally produce negatives of different hardness across denoising stages, substantially increasing sample diversity and boosting knowledge-graph completion performance, it still has several shortcomings \cite{ref12}. \textbf{Firstly, no adaptive mechanism for learning difficulty.} In complex knowledge graphs, entities differ widely in how hard they are to learn. Triples that involve harder entities typically need harder negatives than those involving easier entities. Existing diffusion frameworks struggle to sense and adapt to these differences, leaving hard entities undertrained while easy entities may be over-reinforced. \textbf{Secondly, possible generation of semantically impoverished negatives.} Step-wise denoising helps reduce false negatives, yet early stages can still yield many semantically irrelevant, low-information negative triples, wasting computational resources. \textbf{ Thirdly, under-utilization of negative-sample hardness during training.} Current pipelines usually mix samples from different denoising stages using fixed ratios or simple thresholds, without dynamically adjusting the easy-to-hard balance as learning progresses. This prevents the model from fully leveraging hard negatives to continually improve its discriminative power.

To address the aforementioned limitations and inspired by the success of diffusion models in various generative tasks, this paper proposes a novel \textbf{D}iffusion-based \textbf{A}daptive \textbf{N}egative \textbf{S}ampling method for \textbf{K}nowledge \textbf{G}raph \textbf{C}ompletion (DANS-KGC). First, a \textbf{difficulty assessment module} (DAM) is introduced to perform fine-grained quantification of entity learning difficulty in complex knowledge graphs. Then, a conditional diffusion model is employed to achieve difficulty-aware \textbf{adaptive negative sampling} (ANS). In the forward noise-injection phase, a differentiated noise scheduling strategy is applied: entities with higher learning difficulty are injected with stronger noise, while those with lower difficulty receive weaker noise. In the denoising generation phase, semantic constraints from the positive samples and neighborhood structural information are integrated to ensure that the generated negatives are semantically related to but distinguishable from the positives. By sampling outputs at various timesteps during denoising, the model naturally acquires a multi-level distribution of negative samples with varying hardness. Finally, a \textbf{dynamic training mechanism} (DTM) is proposed to adjust the ratio of negative samples of different hardness levels in real time based on training progress, continuously enhancing triple representation quality and link prediction performance. Our contributions can be summarized as follows:

\begin{itemize}

\item To our knowledge, this study is the first to argue that negative sampling ought to reflect the varying learning difficulties of individual entities. Accordingly, we develop a \textbf{Difficulty Assessment Module (DAM)} that combines semantic cues with graph‑structural signals to produce a quantitative score for each entity's learning difficulty.

\item We devise an \textbf{Adaptive Negative Sampling framework driven by a diffusion process (ANS)}. In the forward diffusion stage, the noise schedule is modulated by each entity's difficulty score: easily learned entities receive lighter perturbations, whereas harder ones are subjected to stronger corruption. During the reverse denoising stage, semantic and neighborhood constraints steer the process so that the recovered triples remain plausible yet diverge from the original positives. By drawing samples at several points along the denoising trajectory, ANS produces a diverse set of negatives spanning a continuum of hardness levels.

\item To maximize the utility of negatives with different hardness levels, we implement a \textbf{Dynamic Training Mechanism (DTM)} that progressively raises the share of hard negatives according to a preset curriculum. Training starts with mostly easy examples, giving the model a stable foundation, and then incrementally introduces more challenging negatives. This staged exposure sharpens the decision boundary step by step, boosting the model's discriminative power and delivering consistent gains in knowledge‑graph completion performance.

\end{itemize}

\section{Related Works}

In this section, we briefly review the related work, mainly covering knowledge graph completion methods, negative sampling techniques in knowledge graphs, and diffusion models.

\subsection{Knowledge Graph Completion Methods}

Translation‑based models remain mainstream. TransE \cite{ref6} embeds entities and relations in a shared vector space, viewing each relation as a translation from head to tail. Variants such as TransH \cite{ref13}, TransR \cite{ref14}, and TransA \cite{ref15} extend this idea with relation‑specific hyperplanes, mapping matrices, or adaptive metrics to better fit 1‑to‑N, N‑to‑1, and N‑to‑N patterns, yet they still struggle with symmetry, antisymmetry, and compositional relations.
Rotation‑based models interpret relations as rotations in complex or hypercomplex space. Starting with RotatE \cite{ref16}, subsequent works—QuatE \cite{ref17}, Rotat3D \cite{ref18}, and DualE \cite{ref19}—capture symmetric, antisymmetric, inverse, and compositional patterns, alleviating the limitations of translation models.
Semantic‑matching methods evaluate triples via latent semantic similarity. RESCAL \cite{ref20} uses a bilinear form with full‑rank relation matrices; DistMult \cite{ref21}, HolE \cite{ref22}, and ComplEx \cite{ref23} refine this by imposing structural constraints (diagonal matrices, circular convolution, or complex embeddings) to boost completion accuracy.

Most of the above reduce a knowledge graph to isolated triples, overlooking richer graph topology. GNN‑based approaches fill this gap by propagating information along edges: convolutional models like R‑GCN \cite{ref24} and CompGCN \cite{ref25} aggregate neighbor features with relation‑specific kernels, while attention models such as KBGAT \cite{ref26} and KRACL \cite{ref27} weight neighbors adaptively. Both lines exploit multi‑hop context, delivering state‑of‑the‑art results in knowledge graph completion.

\subsection{Negative Sampling Methods}

%Negative sampling is a key technique for training completion models, as it generates negative triples that do not exist in the knowledge graph to form contrastive signals with positive triples. The random replacement strategy replaces the head entity, relation, or tail entity without constraints. While simple and low-cost, it easily produces false negatives or semantically trivial samples, weakening the gradient signal. Heuristic rule-based methods introduce entity type constraints, relation mutual-exclusion rules, and statistical priors to filter candidates and select negative samples based on predefined metrics. However, these methods heavily rely on domain knowledge and exhibit limited generalizability across domains. Adversarial sampling uses generative adversarial networks or self-adversarial mechanisms to dynamically generate harder negative distributions, improving the model's discriminative capacity. Still, they lack explicit controllability over the ``hardness" of negatives. The latest diffusion-based sampling introduces a Markovian denoising process into negative sampling, naturally generating a multi-scale sequence of samples from easy to hard, significantly enriching sample diversity. Nevertheless, current diffusion-based approaches remain at the level of global difficulty modeling and fail to achieve fine-grained, adaptive sampling tailored to the highly variable learning difficulties of entities within the knowledge graph.

Negative sampling (NS) is essential in training graph representation learning models, generating negative triples absent from the knowledge graph (KG) to contrast positive triples and improve model discriminative power. Traditional NS strategies, such as random replacement, are simple and efficient but often produce false negatives or trivial samples, weakening training signals. Heuristic-based methods address these limitations through example popularity, prediction scores, or high-variance sample selection. Bernoulli sampling enhances negative sample quality via Bernoulli distributions \cite{ref13}. Graph-based heuristics, such as SANS \cite{ref28}, select negatives from k-hop neighborhoods; MixGCF \cite{ref29} synthesizes hard negatives by mixing hops and positives; and MCNS \cite{ref30} employs Metropolis-Hastings sampling guided by sub-linear positivity theory. However, these methods typically rely on domain-specific knowledge, limiting their generalizability. Adversarial NS techniques like KBGAN \cite{ref31} and IGAN \cite{ref32} utilize Generative Adversarial Networks (GANs) to dynamically create harder negatives, significantly boosting model discriminative capability. Despite improvements, adversarial and caching methods often lack explicit control over negative sample hardness. Recent diffusion-based NS approaches leverage denoising diffusion probabilistic models (DDPMs), generating negatives via step-wise denoising processes, naturally varying sample hardness and increasing diversity \cite{ref11}. However, existing diffusion methods primarily operate globally, lacking fine-grained, entity-specific adaptive sampling.

\subsection{Diffusion Models}

Denoising diffusion probabilistic models (DDPMs) generate data by adding Gaussian noise via a forward Markov chain and then learning a reverse denoising path, achieving highly diverse, high‑fidelity outputs \cite{ref33}. Adaptations to knowledge graphs include FDM, which treats link prediction as conditional generation \cite{ref34}, DHNS, which applies diffusion‑based hierarchical negative sampling to multimodal KGs \cite{ref35}, and other recent explorations \cite{ref36}. However, existing work cannot tailor negative samples to entity‑specific learning difficulty. We address this by introducing a difficulty‑aware noise schedule that adaptively generates negatives for each entity.

\section{Methodology}

\subsection{Problem Definition and Overview}

A knowledge graph can be defined as $\mathcal{G}=(\mathcal{E},\mathcal{R},\mathcal{T})$, $\mathcal{E}$ and $\mathcal{R}$ denote the sets of entities and relations, and $\mathcal{T}$ is the set of triples in the form $\langle h,r,t \rangle$, where $h,t\in \mathcal{E}$ and $r\in \mathcal{R}$. Given an incomplete triple, such as $\langle h,r,?\rangle$ or $\langle ?,r,t\rangle$, where $h$ and $t$ represent known head entities and tail entities, $r$ is relation. The objective of KGC is predicting the missing head or tail entities. During the training process, the primary objective is to distinguish between positive triples $\langle h,r,t \rangle$ and their corresponding negative triples $\langle h',r,t \rangle$ or $\langle h,r,t'\rangle$. The aim is to enhance the discriminative ability of the model through contrastive learning between positive and negative triples. Where $h'$ and $t'$ are corrupted entities obtained through negative sampling.
\begin{figure*}[t]
\centering
\includegraphics[width=1\textwidth]{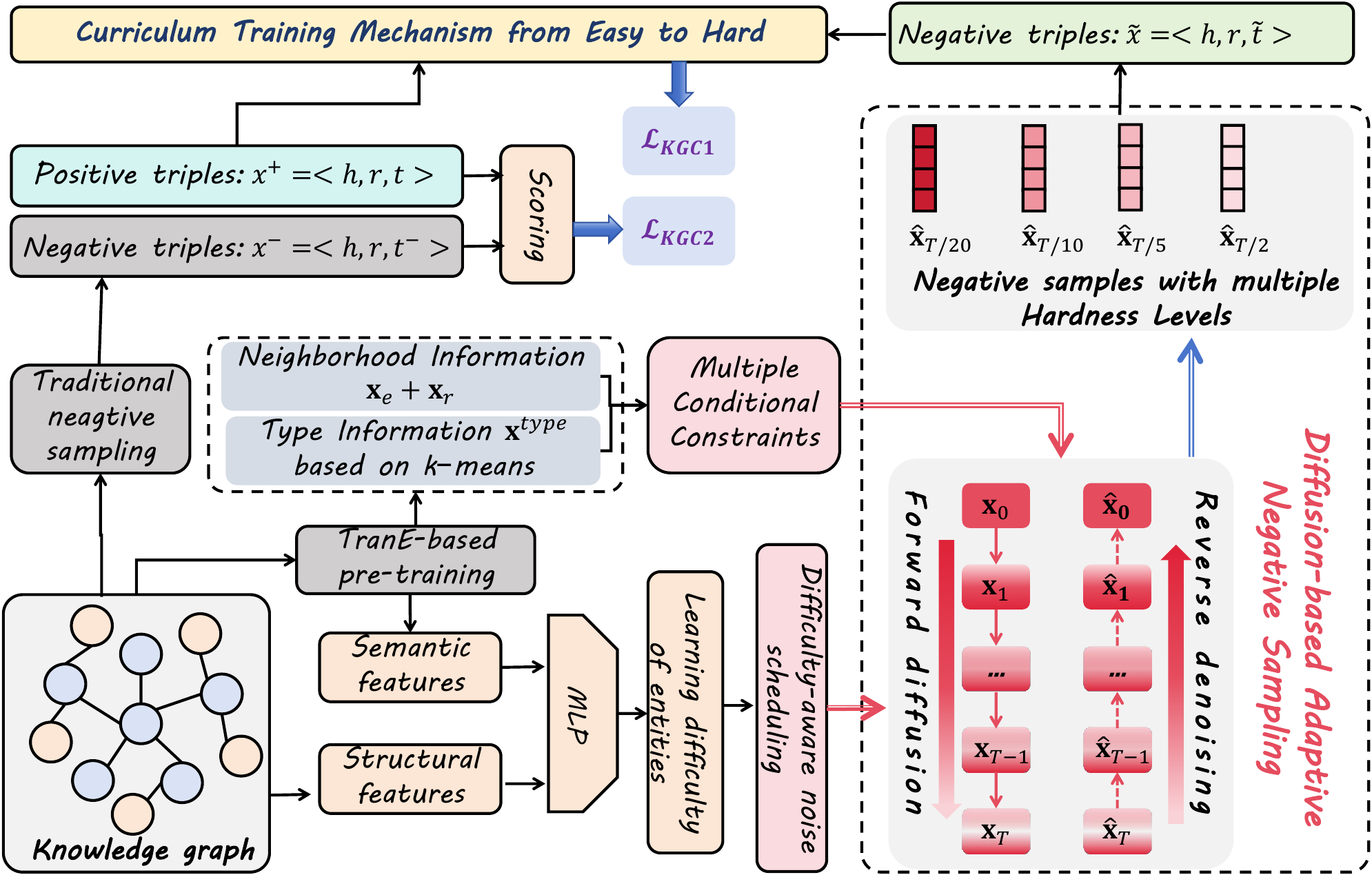} % Reduce the figure size so that it is slightly narrower than the column.
\caption{The overall framework of DANS-KGC.}
\label{fig1}
\end{figure*}

Figure. \ref{fig1} presents the overall architecture of DANS‑KGC, consisting of three cooperating components. The \textbf{Difficulty Assessment Module (DAM)} assigns each entity a difficulty score that drives the subsequent noise scheduling and sampling strategies. The \textbf{Adaptive Negative Sampling Module (ANS)} leverages a diffusion process in which difficulty‑aware noise is injected during the forward pass, and semantic as well as neighborhood constraints are enforced during reverse denoising, producing negatives that are both informative and graph‑consistent. Finally, the \textbf{Dynamic Training Mechanism (DTM)} adjusts the mix of negatives on the fly, steadily shifting from easy to hard according to the curriculum. This staged exposure enables the model to refine its triple representations continuously and achieve superior completion accuracy.

\subsection{Diffculty Assessment Module (DAM)}

As mentioned earlier, due to the complexity of KGs, different entities have varying levels of learning difficulty. We assess each entity's learning difficulty by integrating semantic and structural features of the graph. First, we obtain multi-dimensional features for each entity through pre-training and statistical computations. The semantic features are vector embeddings (e.g., obtained via KG embedding models like TransE), while structural features include metrics such as Betweenness Centrality (BC), Closeness Centrality (CC), Clustering Coefficient (CCoef), Triple Count (TC), Average Neighbor Degree (ADN), and PageRank (PR). Definitions and formulas for these features are provided in Appendix A.

Let $\mathbf{e}_{\text{sem}}$ denote the semantic feature vector of an entity $e$, and let $\mathbf{e}_{\text{str}} = (\mathbf{e}_{\text{str}}^1, \mathbf{e}_{\text{str}}^2, \dots, \mathbf{e}_{\text{str}}^m)$ be its $m$-dimensional structural feature vector. We normalize each feature and then concatenate them to form an entity difficulty representation: $\mathbf{e}^d = [\, \mathbf{e}_{\text{sem}};~\mathbf{e}_{\text{str}}\,]$. Next, a multi-layer perceptron (MLP) maps $e^d$ to a difficulty score:
\begin{equation}
\zeta(e)=sigmoid(\mathbf{W}_2 \cdot ReLU(\mathbf{W}_1 \cdot \mathbf{e}_d+b_1)+b_2)
\end{equation}
\noindent where $\zeta(e) \in (0,1)$ represents the diffculty score of eneity $e$, the closer it is to 0, the lower the learning difficulty of the entity; conversely, the farther it is from 0, the greater the difficulty. $\mathbf{W}_1$ and $\mathbf{W}_2$ represents learnable matrices, respectively. 

\subsection{Diffusion-based Adaptive Negative Sampling (ANS)}

%After obtaining the learning difficulty of each entity in the knowledge graph, we propose an adaptive negative sampling method based on a diffusion model. This method generates appropriate negative samples with varying levels of hardness tailored to entities of different difficulties. By leveraging contrastive learning between positive and negative samples, it enables the learning of higher-quality triple representations. The diffusion model-based adaptive negative sampling consists of two main modules: difficulty-aware forward noise scheduling (DFS) and condition-constrained reverse denoising (CCRD) guided by semantic and neighborhood constraints.

\subsubsection{Difficulty-Aware Forward Noise Scheduling (DFS).}

After deriving each entity’s difficulty score with DAM, we feed this information into the forward noise‑injection stage of the diffusion model: entities judged harder receive stronger perturbations, whereas easier ones are only lightly corrupted. Concretely, for an entity $x$ we set an entity‑specific upper bound on the noise intensity that is proportional to its difficulty $\zeta(x)$:

\begin{equation}
\beta_{max}(x)=\beta_{low}+(\beta_{global}-\beta_{low})(\zeta(x))^{\mu}
\end{equation}
\noindent where $\beta_{max}(x)$ denotes the noise upper bound for sample $x$, $\beta_{\text{global}}$ is a global maximum noise level (set empirically based on prior experience), and $\beta_{low}$ is a minimum noise level applied to all samples to ensure even low-difficulty entities receive some noise (maintaining a baseline level of negative distinguishability). $\mu$ is a hyperparameter controlling the influence of the difficulty score on noise intensity. After assigning difficulty-conditioned noise bounds, we define the noise schedule over time. Let $T$ be the total number of diffusion steps. The noise variance at time step $t$ for entity $x$ is scheduled as:
\begin{equation}
\beta_{t}(x)=\beta_{init}+t/T(\beta_{max}(x)-\beta_{init})
\end{equation}
\noindent where $\beta_{\text{init}}$ is the noise variance at the initial step $t=0$. In this linear schedule, early diffusion steps inject relatively low noise (especially for easy entities with small $\beta_{\max}$), while later steps approach the entity's designated $\beta_{\max}(x)$. Following the DDPM framework, during forward diffusion the input entity embedding $x_0$ is progressively corrupted with noise, eventually yielding $x_T \sim \mathcal{N}(0,I)$ (pure Gaussian noise) after $T$ steps. Formally, the forward diffusion process is a Markov chain:
\begin{equation}
q(x_{1:T} \mid x_0) = \prod_{t=1}^{T} q(\mathbf{x}_t \mid \mathbf{x}_{t-1})
\end{equation}
\begin{equation}
q(\mathbf{x}_t \mid \mathbf{x}_{t-1}) = \mathcal{N}(\mathbf{x}_t; \sqrt{1 - \beta_t} \, \mathbf{x}_{t-1}, \beta_t \mathbf{I})
\end{equation}
where we abbreviate $\beta_t = \beta_t(x)$ for readability. Using the reparameterization trick, one can sample $\mathbf{x}_t$ at any arbitrary step $t$ in closed form:
\begin{equation}
\mathbf{x}_t = \sqrt{\bar{\alpha}_t} \, \mathbf{x}_0 + \sqrt{1 - \bar{\alpha}_t}{\epsilon}_t
\end{equation}
\noindent where $\alpha_t = 1 - \beta_t$, $\bar{\alpha}_t = \prod_{i=1}^{t} \alpha_i$, and $\epsilon_t \sim \mathcal{N}(0, I)$. In this way, we inject noise adaptively according to each entity's difficulty, thereby controlling the hardness of the negative sample that will be generated in the reverse process.

\subsubsection{Condition-Constrained Reverse Denoising (CCD).}

After the forward diffusion process with adaptive noise scheduling, the entity embedding is progressively corrupted, ultimately resulting in a noisy entity representation $\mathbf{x}_t$. The reverse diffusion process iteratively denoises the noisy entity embedding $\mathbf{x}_t$, ultimately obtaining a synthesized entity embedding. Compared to existing reverse denoising processes used for negative sample generation, we impose semantic constraints to ensure the generated triples are semantically valid, thereby enhancing the quality of learning. Given the noisy embedding $\mathbf{x}_t$ at time step $t$, the reverse process is as follows:
\begin{equation}
\begin{split}
p_{\theta}(\mathbf{x}_{t-1} \mid \mathbf{x}_t) = 
&\mathcal{N}(\mathbf{x}_{t-1}; \varphi_{\theta}(\mathbf{x}_t, \mathbf{t}, \mathbf{x}^{\text{type}},\\
 &\mathbf{x}_e+\mathbf{x}_r), \beta_t \mathbf{I})
\end{split}
\label{eq1}
\end{equation}
\noindent where $\theta$ denotes the parameterized configuration of the diffusion model, and $\mathbf{t}$ represents the embedding of time step $t$; $\mathbf{x}_e+\mathbf{x}_r$ is the combined representation of the observed entity $e$ and its corresponding relation $r$, which guides the generation of the missing entity to form a negative sample triple. $\mathbf{x}^{\text{type}}$ represents the semantic type embedding of the observed entity, which is used to impose semantic constraints on the generated negative samples. This ensures that the generated samples maintain a certain degree of semantic relevance to the observed sample, preventing the generation of implausible or meaningless triples. We obtain $x_{\text{type}}$ by performing K-means clustering on the pre-trained entity embeddings and using the representation of each cluster center as the semantic type embedding for all entities within that cluster. Specifically, $\varphi_\theta(\mathbf{x}_t, \mathbf{t}, \mathbf{x}_e, \mathbf{x}_r)$ denotes the mean of the Gaussian distribution, which is defined as:
\begin{equation}
\begin{split}
&\varphi_{\theta}(\mathbf{x}_t, \mathbf{t}, \mathbf{x}^{\text{type}}, \mathbf{x}_e+\mathbf{x}_r) =\\ 
&\frac{1}{\sqrt{\alpha_t}} \mathbf{x}_t - \frac{1 - \sqrt{\alpha_t}}{\sqrt{\alpha_t} \sqrt{1 - \beta_t}} 
\epsilon_{\theta}(\mathbf{x}_t, \mathbf{t}, \mathbf{x}^{\text{type}}, \mathbf{x}_e+\mathbf{x}_r)
\end{split}
\label{eq2}
\end{equation}
After predicting the corresponding noise, the negative sample at the current time step can be obtained by removing the noise from the entity embedding at that time. The formula for generating the corresponding negative sample is as follows:
\begin{equation}
\hat{\mathbf{x}}_{t-1} = \frac{1}{\sqrt{\alpha_t}} \hat{\mathbf{x}}_t - \frac{1 - \sqrt{\alpha_t}}{\sqrt{\alpha_t} \sqrt{1 - \alpha_t}}\epsilon_{\theta}(\mathbf{x}_t, \mathbf{t}, \mathbf{x}^{\text{type}}, \mathbf{x}_e+\mathbf{x}_r) + \sqrt{\beta_t} \epsilon_t
\end{equation}
\noindent where $\hat {\mathbf{x}}_T = \mathbf{x}_T$ denotes the noisy entity embedding at the final time step $T$, and $\mathbf{x}_{t-1}$ represents the intermediate entity embedding at time step $t-1$ within the range $[0, T-1]$. To enhance the diversity of generated samples, $\sqrt{\beta_t} \epsilon_t$ is used to introduce random noise at each step, resulting in intermediate denoised outputs at each iteration. We use a multilayer perceptron combined with a layer normalization layer to predict the noise using learnable parameters $\theta$ as follows:
\begin{equation}
\epsilon_{\theta}(\mathbf{x}_t, \mathbf{t}, \mathbf{x}^{\text{type}}, \mathbf{x}_e+\mathbf{x}_r)=LayerNorm(MLP(\mathbf{x}_t, \mathbf{t}, \mathbf{x}^{\text{type}}))
\end{equation}
We create negative samples by corrupting the tail entities of positive sample triples. Specifically, we feed the head entity, entity type, neighborhood, and temporal embeddings as conditional inputs to the reverse denoising function to generate semantically rich negative triples. To optimize this process, we minimize the denoising diffusion loss function to generate high-quality embeddings for the corrupted tail entities.
\begin{equation}
\mathcal{L}_{Diff}=||\epsilon_{\theta}(\mathbf{x}_t, \mathbf{t}, \mathbf{x}^{\text{type}}, \mathbf{x}_e+\mathbf{x}_r)-\epsilon_{t}||^2
\end{equation}
\noindent which is the mean squared error between the predicted noise and the true noise at each diffusion step (with $\epsilon_t$ from the forward process). By training with $\mathcal{L}_{\text{Diff}}$, the model learns to generate high-quality negative entity embeddings that correspond to realistic corruptions of the original triple.

As we sample the reverse process at different intermediate time steps, the resulting negative samples exhibit different levels of difficulty: negatives obtained after fewer denoising steps (small $t$) are closer to the original positive and thus harder to distinguish, whereas those from later steps (large $t$) are more distorted by noise and thus easier. Therefore, by selecting particular diffusion steps, we can generate negative samples of varying difficulty for each positive triple. In this paper, we sample negatives at four diffusion steps (from near the middle to near the end of the process): $t \in \{\,T/20,\;T/10,\;T/5,\;T/2\,\}$. This yields a set of negative triples with graded difficulty:
\begin{equation}
G^- = \{\; \langle h, r, \hat{x}_t \rangle \mid t \in \{T/20,\;T/10,\;T/5,\;T/2\} \;\},
\end{equation}
\noindent where $\hat{x}_t$ denotes the generated tail entity embedding at reverse step $t$ (i.e., using $\hat{x}_t$ instead of continuing to $\hat{x}_0$). 

In summary, the ANS module adaptively generates negative samples of varying hardness for entities of different difficulties via difficulty-based noise scheduling, conditional constrained denoising, and hierarchical sampling from multiple diffusion timesteps. Compared to simple random negative generation, this is a more refined strategy that provides a richer set of negatives. By integrating these diverse negatives into training, we can supply stronger and more informative training signals to the KGC model.

\subsection{Dynamic Training Mechanism (DTM)}

To effectively leverage the negative samples of varying difficulty produced by ANS, we propose a curriculum-style dynamic training mechanism (DTM). DTM gradually shifts training focus from easy negatives to hard negatives as learning progresses, stabilizing early training and progressively challenging the model with harder examples.

\subsubsection{Difficulty-based partitioning.}To fully exploit the difficulty‑aware negatives produced at the four pre‑defined diffusion timesteps $\mathcal{T}=\{\tfrac{T}{20},\tfrac{T}{10},\tfrac{T}{5},\tfrac{T}{2}\}$,we redesign the curriculum schedule so that every difficulty band \emph{exactly} corresponds to one sampling timestep. Let $t(\tilde{x})$ denote the diffusion timestep used to generate negative triple $\tilde{x}\!\in\!G^{-}$.We group $G^{-}$ into four disjoint subsets:
\begin{equation}
\mathcal{N}^{(k)}=\bigl\{\tilde{x}\in G^{-}\ \big|\ t(\tilde{x})=\mathcal{T}_{k}\bigr\},
\quad k=1,2,3,4,
\end{equation}
\noindent where $\mathcal{T}_{1}<\mathcal{T}_{2}<\mathcal{T}_{3}<\mathcal{T}_{4}$ and the difficulty monotonically increases with $k$. Smaller timesteps preserve more noise, making the generated negatives harder; hence $\mathcal{T}_{1}=\tfrac{T}{20}$ is the hardest band.
%we denote $t(\tilde{x})$ as the denoising timestep associated with each negative triple $\tilde{x}\in\mathcal{G}^{-}$. We partition $[1,T]$ into four intervals with increasing difficulties and accordingly define four negative subsets:
%\begin{equation}
%\begin{aligned}
%\mathcal{I}_1&=[1,\tfrac{T}{20}],\quad
%\mathcal{I}_2=[\tfrac{T}{20},\tfrac{T}{10}],\\[3pt]
%\mathcal{I}_3&=[\tfrac{T}{10},\tfrac{T}{5}],\quad
%\mathcal{I}_4=[\tfrac{T}{5},\tfrac{T}{2}],
%\end{aligned}
%\end{equation}
%\begin{equation}
%\begin{split}
%\mathcal{N}^{(k)}=\{\tilde{x}\in\mathcal{G}^{-}\mid t(\tilde{x})\in\mathcal{I}_k\},\quad k=1,2,\dots,4\\
%\end{split}
%\end{equation}
%
%The difficulty of the negative samples in $\mathcal{N}^{(1)}$ to $\mathcal{N}^{(4)}$  increases progressively.

\subsubsection{Curriculum sampling schedule.}
We denote the training epoch as $e$, and the total number of epochs as $E_{max}$, with the normalized training progress defined as $\tau_e=e/E_{max}\in[0,1]$. Given the stage boundaries $(b_0,b_1,b_2,b_3,b_4)=(0,0.25,0.5,0.75,1)$, the sampling weight for difficulty band $k$ at epoch $e$ is:
\begin{equation}
w_e^{(k)}=\frac{\exp\bigl(\lambda\,[\tau_e-b_{k-1}]_{+}^{\zeta}\bigr)}
{\sum_{j=1}^{4}\exp\bigl(\lambda\,[\tau_e-b_{j-1}]_{+}^{\zeta}\bigr)},
\quad k=1,\dots,4,
\end{equation}
\noindent where $[x]_+=\max(0,x)$, $\zeta\in[0,1]$ controls smoothness, and $\lambda>0$ sharpens or softens the distribution.

\subsubsection{ Mini-batch Training.}
For each positive triple $x^{+}=\langle h,r,t\rangle$, we sample a difficulty band $\kappa\sim\mathrm{Cat}(w_e^{(1:4)})$ based on the current weights, and then select an entity from $\mathcal{N}^{(\kappa)}$ to corrupt the head or tail to obtain a negative triple $\tilde{x}$. Then we define the stage-aware weighted margin loss as follows:
%\begin{equation}
%\mathcal{L}_e=\frac{1}{|\mathcal{B}_e|}
%\sum_{(x^{+},\tilde{x})\in\mathcal{B}_e}
%w_e^{(\kappa(\tilde{x}))}
%\max\bigl(0,\;m_{\kappa}(\tau_e)+f_{\theta}(\tilde{x})-f_{\theta}(x^{+})\bigr)
%\end{equation}
\begin{equation}
\begin{split}
\mathcal{L}_{KGC1}=&-log\sigma(\gamma_{\kappa}(\tau_e)-S(x^{+}))\\
&-\frac{1}{|\mathcal{B}_e|}\sum_{\tilde{x}\in \mathcal{B}_e}w_e^{(\kappa(\tilde{x}))}log\sigma(S(\tilde{x})-\gamma_{\kappa}(\tau_e))
\end{split}
\end{equation}
\noindent where
\begin{equation}
\gamma_{\kappa}(\tau_e)=\gamma_{base}\cdot\bigl[1+\beta\cdot\mathrm{hard}_{\kappa}\cdot\tau_e\bigr],\quad \mathrm{hard}_{\kappa}=\frac{5-\kappa}{4},
\end{equation}
and $\gamma_{base}$ is the base margin, $\beta\in[0,1]$ controls the dynamic margin increment, and $S(\cdot)$ is the triple scoring function. As training proceeds, higher-difficulty weights and margins progressively dominate the loss calculation, focusing gradient updates on harder negatives. In addition, we also retained the negative triples obtained by negative sampling based on random replacement, which are also used to train knowledge graph representations, with the loss function being:
\begin{equation}
\begin{split}
\mathcal{L}_{KGC2}=&-log\sigma(\gamma_{base}-S(x^{+}))\\
&-\frac{1}{|\mathcal{B}^{-}|}\sum_{x^{-}\in \mathcal{B}^{-}}log\sigma(S(x^{-})-\gamma_{base})
\end{split}
\end{equation}
\noindent where $\gamma_{base}$ indicates the fixed margin, $x^{-}$ represents the negative sample from random replacement techniques. The total loss for KGC is as follows:
\begin{equation}
\mathcal{L}_{KGC}=\eta \mathcal{L}_{KGC1}+\mathcal{L}_{KGC2}
\end{equation}
Furthermore, the loss function of the overall algorithm is as follows:
\begin{equation}
\mathcal{L}=\mathcal{L}_{KGC}+\mathcal{L}_{Diff}
\end{equation}
To facilitate the understanding of the entire algorithm, Algorithm \ref{alg:training} presents the overall process of DANS-KGC.
\begin{algorithm}[h]
\small
\caption{DANS-KGC}
\label{alg:training}
\begin{algorithmic}[1]
\STATE \textbf{Input:} Training triples $T_{\text{train}}$; difficulty features for entities (from DAM); total diffusion steps $T$.
\STATE Compute initial difficulty scores $\zeta(e)$ for all entities $e$ using DAM.
\FOR{epoch $ep = 1$ \textbf{to} $E_{\max}$}
    \STATE Update curriculum weights $w_{ep}^{(1:4)}$ based on $\tau_{ep} = ep/E_{\max}$.
    \FOR{each positive triple $x^+ = \langle h, r, t \rangle$ in a batch $B^+$}
        \STATE Sample band $\kappa \sim \text{Categorical}(w_{ep}^{(1:4)})$.
        \STATE Generate $G^- = \{\langle h, r, \hat{x}_t \rangle| t = T/20, T/10, T/5, T/2\}$ using ANS (forward diffusion with Eq. 2--3, reverse diffusion with Eq.7--9).
        \STATE Select a negative $\tilde{x}$ from $N^{(\kappa)} \subset G^-$ for $x^+$.
        \STATE Sample a random negative $x^- = \langle h, r, t' \rangle$ (or $\langle h', r, t \rangle$) by corrupting $x^+$.
    \ENDFOR
    \STATE Compute $\mathcal{L}_{KGC1}$ on batch $B^+$ and corresponding negatives $\{\tilde{x}\}$ using Eq. 15.
    \STATE Compute $\mathcal{L}_{KGC1}$ on batch $B^+$ and random negatives $\{x^-\}$.
    \STATE Compute diffusion loss $\mathcal{L}_{\text{Diff}}$ on generated negatives.
    \STATE Update model parameters by minimizing $\mathcal{L} = \eta \mathcal{L}_{KGC1} + \mathcal{L}_{KGC2} + \mathcal{L}_{\text{Diff}}$.
\ENDFOR
\end{algorithmic}
\end{algorithm}

\section{Experiments}

In this section, to illustrate the superiority of DNS-KGC, we conducted comprehensive experiments on the link prediction task.

\subsection{Experiment Setting}

\subsubsection{Datasets:}We evaluate on six public KGC datasets: \textbf{Family}, \textbf{UMLS}, \textbf{WN18RR}, \textbf{FB15K-237}, \textbf{NELL-995}, and \textbf{YAGO3-10}. Statistics of each dataset are provided in the Table \ref{dataset}.

\begin{table}[ht]
\centering
\begin{tabular}{lcccccc}
\toprule
\textbf{Dataset} & \textbf{Entity} & \textbf{Relation}  & \textbf{Train} &  \textbf{Valid} & $ \textbf{Test}$ \\
\midrule
Family     & 3.0k   & 12     & 5.9k   & 2.0k   & 2.8k \\
UMLS       & 135    & 46       & 1.3k   & 569    & 633  \\
WN18RR     & 40.9k  & 11     & 21.7k  & 3.0k   & 3.1k \\
FB15k-237   & 14.5k  & 237    & 68.0k  & 17.5k  & 20.4k \\
NELL-995   & 74.5k  & 200    & 37.4k  & 543    & 2.8k \\
YAGO3-10   & 123.1k & 37     & 269.7k & 5.0k   & 5.0k \\
\bottomrule
\end{tabular}
\caption{Statistics of datasets used in experiments. Train, Valid, and Test represent the number of training, validation, and test queries, respectively.}
\label{dataset}
\end{table}

\subsubsection{Baseline:}To comprehensively evaluate the effectiveness of DNS-KGC, we compared it with the following three categories of methods. \textbf{Traditional embedding methods}: ConvE \cite{ref38}, QuatE \cite{ref17}, RotatE \cite{ref16}, DRUM \cite{ref39}, and RNNLogic \cite{ref40}; \textbf{Graph neural network-based methods}: CompGCN, NBFNet \cite{ref41}, ConGLR \cite{ref42}, and RED-GNN \cite{ref43}; \textbf{Diffusion-based generative methods}: FDM \cite{ref34} and DiffusionE \cite{ref10}.

\subsubsection{Evaluation Protocols:}During the testing process, for each triple $\langle h, r, t \rangle$, we construct two queries $\langle h, r, ? \rangle$ and $\langle ?, r, t \rangle$, whose answers are $t$ and $h$ respectively. We then calculate the scores of the candidate triples and rank the candidate entities in descending order based on the scores. The evaluation metrics used are Mean Reciprocal Rank (MRR) and Hit@N (N=1,10). Higher MRR and Hit@N values indicate better performance.

\subsubsection{Implementation Details:}Our model is implemented in PyTorch and trained on four NVIDIA A6000 GPUs. We use the Adam optimizer. The hyperparameters (learning rate, diffusion steps $T$, noise hyperparameters $\beta_{\text{global}}, \beta_{\text{low}}, \mu$, curriculum parameters $\lambda, \zeta, \beta$, etc.) are tuned on validation sets; their chosen values are provided in the Table \ref{tab:hyperparameters}.

\begin{table*}[h]
\centering
\begin{tabular}{lcccccc}
\hline
\textbf{Hyperparameter} & \textbf{Family} & \textbf{UMLS} & \textbf{WN18RR} & \textbf{FB15K-237} & \textbf{NELL-995} & \textbf{YAGO3-10} \\ \hline
Learning rate         & 8$e^{-5}$            & 5$e^{-5}$          & 3$e^{-4}$            & 3$e^{-4}$               & 5$e^{-4}$              & $e^{-3}$               \\
Batchsize             & 256             & 256           & 512             & 512                & 1024              & 2048              \\
Epoch                 & 1500            & 1500          & 1000            & 2000               & 2000              & 2500              \\
Embedding size        & 200             & 200           & 400             & 500                & 500               & 600               \\
$\lambda$             & 10              & 10            & 10              & 5                  & 5                 & 5                 \\
$\zeta$               & 1               & 1             & 1               & 0.75               & 1                 & 0.75              \\
$\eta$                & 0.30            & 0.40          & 0.20            & 0.25               & 0.25              & 0.25              \\
$\gamma$              & 1               & 1             & 1               & 1                  & 1                 & 1                 \\
$\mu$                 & 1               & 1             & 1.5             & 2                  & 2                 & 2                 \\
Optimizer             & AdamW           & AdamW         & AdamW           & AdamW              & AdamW             & AdamW             \\
$\beta$               & 0.4             & 0.4           & 0.4             & 0.4                & 0.4               & 0.3               \\ \hline
\end{tabular}
\caption{Hyperparameter settings for different datasets.}
\label{tab:hyperparameters}
\end{table*}

\subsection{Comparison Experiment Results}

\begin{table*}[t]
\centering
\begin{tabular}{l|ccc|ccc|ccc}
\toprule

\multirow{2}{*}{\textbf{Models}} & \multicolumn{3}{c|}{\textbf{Family}} & \multicolumn{3}{c|}{\textbf{UMLS}} & \multicolumn{3}{c}{\textbf{WN18RR}} \\

 &\textbf{MRR} & \textbf{H@1} & \textbf{H@10} & \textbf{MRR} & \textbf{H@1} & \textbf{H@10} & \textbf{MRR} & \textbf{H@1} & \textbf{H@10} \\
\midrule
ConvE      & 0.912 & 0.837 & 0.982 & 0.937 & 0.922 & 0.967 & 0.427 & 0.392 & 0.498 \\
QuatE      & 0.941 & 0.896 & 0.991 & 0.944 & 0.905 & 0.993 & 0.480 & 0.440 & 0.551 \\
RotatE     & 0.921 & 0.866 & 0.988 & 0.925 & 0.863 & 0.993 & 0.477 & 0.428 & 0.571 \\
DRUM       & 0.934 & 0.881 & 0.996 & 0.813 & 0.674 & 0.976 & 0.486 & 0.425 & 0.586 \\
RNNLogic   & 0.881 & 0.857 & 0.907 & 0.842 & 0.772 & 0.965 & 0.483 & 0.446 & 0.558 \\
CompGCN    & 0.933 & 0.883 & 0.991 & 0.927 & 0.867 & \underline{0.994} & 0.479 & 0.443 & 0.546 \\
NBFNet     & 0.989 & 0.988 & 0.989 & 0.948 & 0.920 & \textbf{0.995} & 0.551 & 0.497 & \textbf{0.666} \\
RED-GNN    & \underline{0.992} & \underline{0.988} & \textbf{0.997} & 0.964 & 0.946 & 0.990 & 0.533 & 0.485 & 0.624 \\
FDM        & - & - & - & 0.922 & 0.893 & 0.970 & 0.506 & 0.456 & 0.592 \\
DiffusionE & 0.990 & 0.989 & 0.992 & \underline{0.970} & \underline{0.957} & 0.992 & \underline{0.557} & \underline{0.504} & 0.658 \\
\rowcolor{myred}
DANS-KGC   & \textbf{0.994} & \textbf{0.991} & \underline{0.995} & \textbf{0.972} & \textbf{0.962} & \textbf{0.995} & \textbf{0.558} & \textbf{0.509} & \underline{0.661} \\
\bottomrule
\end{tabular}
\caption{Comparison on Family, UMLS, and WN18RR. Best results are in \textbf{bold}, second-best in \underline{underline}.}
\label{tab1}
\end{table*}

\begin{table*}[t]
\centering
\begin{tabular}{l|ccc|ccc|ccc}
\toprule

\multirow{2}{*}{\textbf{Models}} & \multicolumn{3}{c|}{\textbf{FB15K-237}} & \multicolumn{3}{c|}{\textbf{NELL-995}} & \multicolumn{3}{c}{\textbf{YAGO3-10}} \\

& \textbf{MRR} & \textbf{H@1} & \textbf{H@10} & \textbf{MRR} & \textbf{H@1} & \textbf{H@10} & \textbf{MRR} & \textbf{H@1} & \textbf{H@10} \\
\midrule
ConvE      & 0.325 & 0.237 & 0.501 & 0.511 & 0.446 & 0.619 & 0.520 & 0.450 & 0.660 \\
QuatE      & 0.350 & 0.256 & 0.538 & 0.533 & 0.466 & 0.643 & 0.379 & 0.301 & 0.534 \\
RotatE     & 0.337 & 0.241 & 0.533 & 0.508 & 0.448 & 0.608 & 0.495 & 0.402 & 0.670 \\
DRUM       & 0.343 & 0.255 & 0.516 & 0.532 & 0.460 & \underline{0.662} & 0.531 & 0.453 & 0.676 \\
RNNLogic   & 0.344 & 0.252 & 0.530 & 0.416 & 0.363 & 0.478 & 0.554 & \underline{0.509} & 0.622 \\
CompGCN    & 0.355 & 0.264 & 0.535 & 0.463 & 0.383 & 0.596 & 0.421 & 0.392 & 0.577 \\
NBFNet     & \underline{0.415} & 0.321 & \underline{0.599} & 0.525 & 0.451 & 0.639 & 0.550 & 0.479 & 0.686 \\
RED-GNN    & 0.374 & 0.283 & 0.558 & 0.543 & 0.476 & 0.651 & 0.559 & 0.483 & 0.689 \\
FDM        & \textbf{0.485} & \textbf{0.386} & \textbf{0.681} & - & - & - & - & - & - \\
DiffusionE & 0.376 & 0.294 & 0.539 & \textbf{0.552} & \underline{0.490} & 0.654 & \underline{0.566} & 0.494 & \underline{0.692} \\
\rowcolor{myred}
DANS-KGC   & 0.404 & \underline{0.324} & 0.596 & \underline{0.546} & \textbf{0.512} & \textbf{0.669} & \textbf{0.572} & \textbf{0.512} & \textbf{0.701} \\
\bottomrule
\end{tabular}
\caption{Comparison on FB15K-237, NELL-995, and YAGO3-10. Best results are in \textbf{bold}, second-best in \underline{underline}.}
\label{tab2}
\end{table*}

As shown in Table~\ref{tab1} and Table~\ref{tab2}, \textbf{DANS-KGC achieves superior performance across all six benchmark datasets}. On \textbf{UMLS} and \textbf{YAGO3-10}, our method achieves the highest scores on all three metrics, with MRRs of 0.972 and 0.572 respectively, demonstrating strong modeling and generalization capability. On \textbf{WN18RR}, DANS-KGC also obtains the best MRR (0.558), outperforming all baselines including recent diffusion-based methods. For the \textbf{Family} dataset, DANS-KGC reaches an MRR of 0.994 and Hits@1 of 0.991, confirming its excellent reasoning ability on this domain. On \textbf{FB15K-237} and \textbf{NELL-995}, our model ranks among the top performers, with an MRR of 0.546 on NELL-995 and the highest Hits@10 (0.669) on NELL-995. These results highlight the effectiveness of our adaptive sampling and dynamic training mechanisms in improving link prediction across diverse scenarios.

\subsection{Abaltion Study}

We perform ablation experiments on the \textbf{Family} dataset to quantify the contribution of each proposed component. Specifically, we consider variants: \textbf{DANS w/o DFS}, which uses a simple linear noise schedule (removing the difficulty-based forward scheduling); \textbf{DANS w/o CCD}, which removes the entity type and neighborhood semantic constraints during reverse diffusion (the diffusion model generates negatives without conditional guidance); and \textbf{DANS w/o DTM}, which replaces the curriculum-based dynamic training with a traditional static training regime (e.g., mixing easy and hard negatives in fixed proportion). Specific ablation study results and analysis are presented in the Table \ref{ablation}.

\begin{table}[h]
\centering
\begin{tabular}{lccc}
\toprule
\textbf{Model} & \textbf{MRR} & \textbf{Hit@1} & \textbf{Hit@10} \\
\midrule
DANS-KGC         & 0.994 & 0.991 & 0.995 \\
DANS w/o DFS     & 0.965 & 0.943 & 0.988 \\
DANS w/o CCD     & 0.989 & 0.976 & 0.983 \\
DANS w/o DTM     & 0.978 & 0.921 & 0.974 \\
\bottomrule
\end{tabular}
\caption{Ablation study of DANS-KGC on Family dataset.}
\label{ablation}
\end{table}

\textbf{Difficulty-based forward noise scheduling.} Removing the DFS module results in a significant performance drop, indicating that adapting the noise intensity according to entity difficulty is essential for generating appropriately hard negative samples. This mechanism ensures that entities with higher learning difficulty are exposed to stronger perturbations, enabling the model to better distinguish them from similar negatives.

\textbf{Condition-constrained reverse denoising.} The absence of the CCD module also leads to a noticeable decline in performance, which confirms the importance of incorporating semantic type and neighborhood constraints in the denoising process. These constraints guide the generation of semantically plausible but still challenging negative triples, enhancing both the quality and informativeness of the training signals.

\textbf{Dynamic training mechanism.} Replacing the DTM module with a static training process results in a marked reduction in Hits@1, underscoring the value of curriculum-style training. The dynamic adjustment of negative sample difficulty over training epochs helps the model to first stabilize with easier examples and then gradually focus on harder ones, thereby achieving more robust optimization and improved generalization.

\subsection{Hyperparameter Senstivity Analysis}

To comprehensively evaluate the performance of the DANS-KGC model under different configurations, we conducted sensitivity experiments on two key hyperparameters—$\mu$ and $\eta$. These parameters govern the model's core mechanisms and significantly impact its final performance.

First, $\mu$ controls the intensity of adaptive noise scheduling. It determines how the model generates challenging negative samples based on the learning difficulty of entities. By adjusting the value of $\mu$, we aim to explore the model's reliance on this adaptive negative sampling strategy. Understanding the sensitivity of $\mu$ will help reveal the effectiveness of injecting stronger noise into harder entities and its contribution to the model's overall performance. Second, $\eta$ balances the loss between DANS negative samples and traditional random negative samples. This parameter directly reflects the model’s dependence on the proposed DANS sampling mechanism. By varying $\eta$, we can analyze how the DANS mechanism influences the training process and final performance under different weight settings. In particular, when $\eta$ approaches zero, the model primarily relies on traditional random negative sampling, which provides strong evidence for the irreplaceability of the DANS mechanism.

\begin{figure}[ht]
\centering
\includegraphics[width=1\columnwidth]{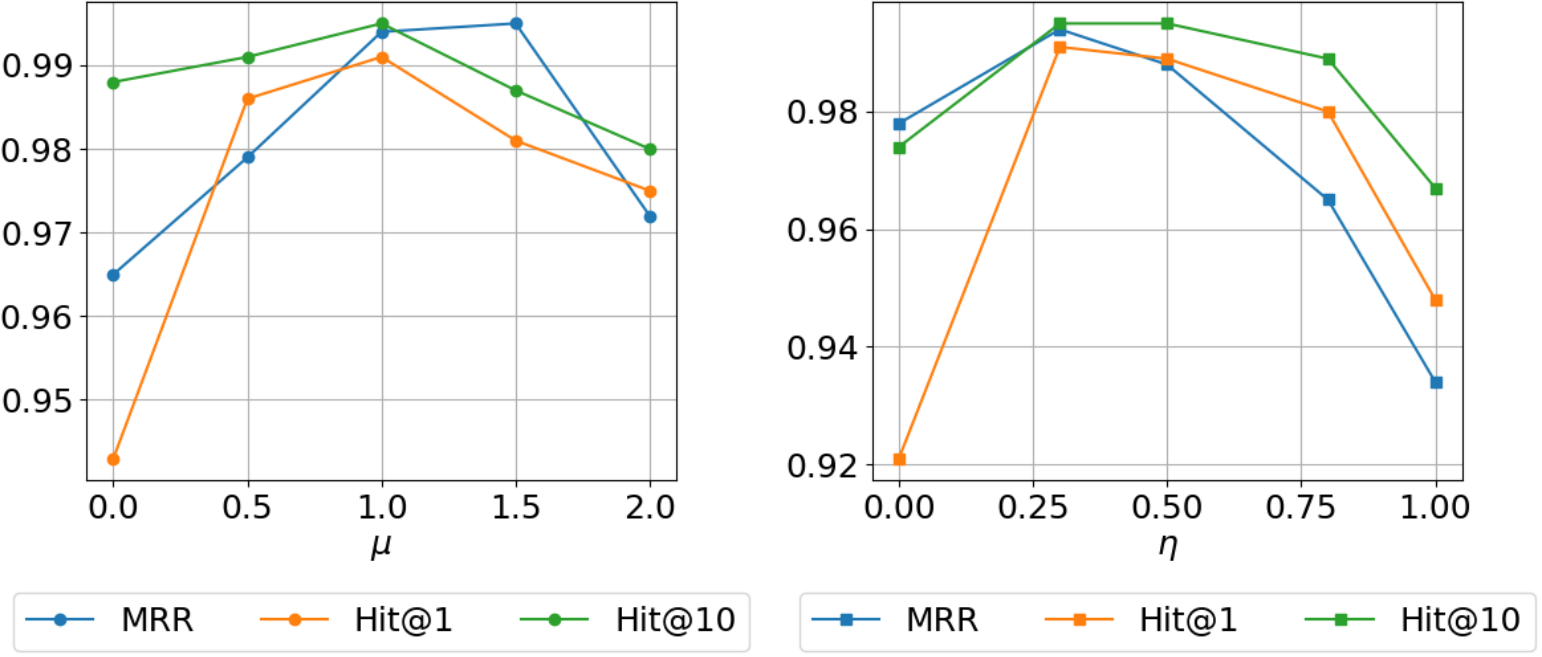} % Reduce the figure size so that it is slightly narrower than the column. Don't use precise values for figure width.This setup will avoid overfull boxes.
\caption{The sensitivity analysis results of parameters $\mu$ and $\eta$ on the Family dataset.}
\label{fig1}
\end{figure}

The model shows notable sensitivity to the hyperparameters $\mu$ and $\eta$. Specifically, $\mu$ controls the strength of adaptive noise scheduling. The best performance (MRR, Hit@1, Hit@10) is achieved when $\mu = 1$, indicating its effectiveness in generating challenging negative samples. However, extreme values (e.g., $\mu = 0.0$ or $2.0$) lead to performance degradation, suggesting the necessity of properly tuning noise intensity. The parameter $\eta$ balances the loss between DANS-generated and randomly sampled negatives. The optimal performance occurs at $\eta = 0.25$, highlighting the effectiveness of the DANS mechanism. Performance drops significantly as $\eta \to 0.0$, reflecting over-reliance on random sampling, while excessive values of $\eta$ also impair performance, implying the need for a balanced sampling strategy.

\section{Conclusions}

To address the issue that existing methods struggle to achieve adaptive negative sampling for different entities, this paper proposes the DANS-KGC method. This method introduces a noise scheduling mechanism based on difficulty and a denoising module with conditional constraints, utilizing diffusion models to generate diverse negative samples for entities of different difficulties. Moreover, a dynamic training mechanism is introduced during the training process to gradually make full use of various negative samples. Experimental results show that DANS-KGC achieves excellent performance on multiple datasets.

\bibliography{aaai2026}

\end{document}